\pdfoutput=1

\documentclass[11pt]{article}

\usepackage[]{acl}

\usepackage{times}
\usepackage{enumitem}
\usepackage{amsmath}
\usepackage{amsfonts}
\usepackage{latexsym}
\usepackage{graphicx}
\usepackage{makecell}
\usepackage{multirow}
\usepackage{booktabs}

\usepackage[T1]{fontenc}

\usepackage[utf8]{inputenc}

\usepackage{microtype}

\title{UniEX: An Effective and Efficient Framework for Unified Information Extraction via a Span-extractive Perspective}

\author{
    Ping Yang$^\textnormal{1}$\footnotemark[1] \qquad
    Junyu Lu$^\textnormal{12}$\footnotemark[1] \qquad
    Ruyi Gan$^\textnormal{1}$\footnotemark[1] \qquad
    Junjie Wang$^\textnormal{3}$ \qquad
    Yuxiang Zhang$^\textnormal{3}$ \qquad
    \\
    \textbf{
    Jiaxing Zhang$^\textnormal{1}$\footnotemark[2] \qquad
    Pingjian Zhang$^\textnormal{2}$\footnotemark[2] \qquad
    }
    \\
    $^\textnormal{1}$International Digital Economy Academy \quad
    $^\textnormal{2}$South China University of Technology \quad \\
    $^\textnormal{3}$Waseda University \\
    {\tt\small \{yangping, lujunyu, ganruyi\}@idea.edu.cn } \\
    {\tt\small wjj1020181822@toki.waseda.jp, joel0495@asagi.waseda.jp} \\ 
    {\tt\small pjzhang@scut.edu.cn} \quad 
}

\begin{document}
\maketitle

{
  \renewcommand{\thefootnote}%
    {\fnsymbol{footnote}}
  \footnotetext[1]{Equal Contribution.}
  \footnotetext[2]{Corresponding Author.}
}

\begin{abstract}

We propose a new paradigm for universal information extraction (IE) that is compatible with any schema format and applicable to a list of IE tasks, such as named entity recognition, relation extraction, event extraction and sentiment analysis. Our approach converts the text-based IE tasks as the token-pair problem, which uniformly disassembles all extraction targets into joint span detection, classification and association problems with a unified extractive framework, namely UniEX. UniEX can synchronously encode schema-based prompt and textual information, and collaboratively learn the generalized knowledge from pre-defined information using the auto-encoder language models. We develop a traffine attention mechanism to integrate heterogeneous factors including tasks, labels and inside tokens, and obtain the extraction target via a scoring matrix. Experiment results show that UniEX can outperform generative universal IE models in terms of performance and inference-speed on $14$ benchmarks IE datasets with the supervised setting. The state-of-the-art performance in low-resource scenarios also verifies the transferability and effectiveness of UniEX.

\end{abstract}

\section{Introduction}
\label{sec:introduction}

\begin{figure}[t]
\begin{center}
\centerline{\includegraphics[width=1.0 \linewidth]{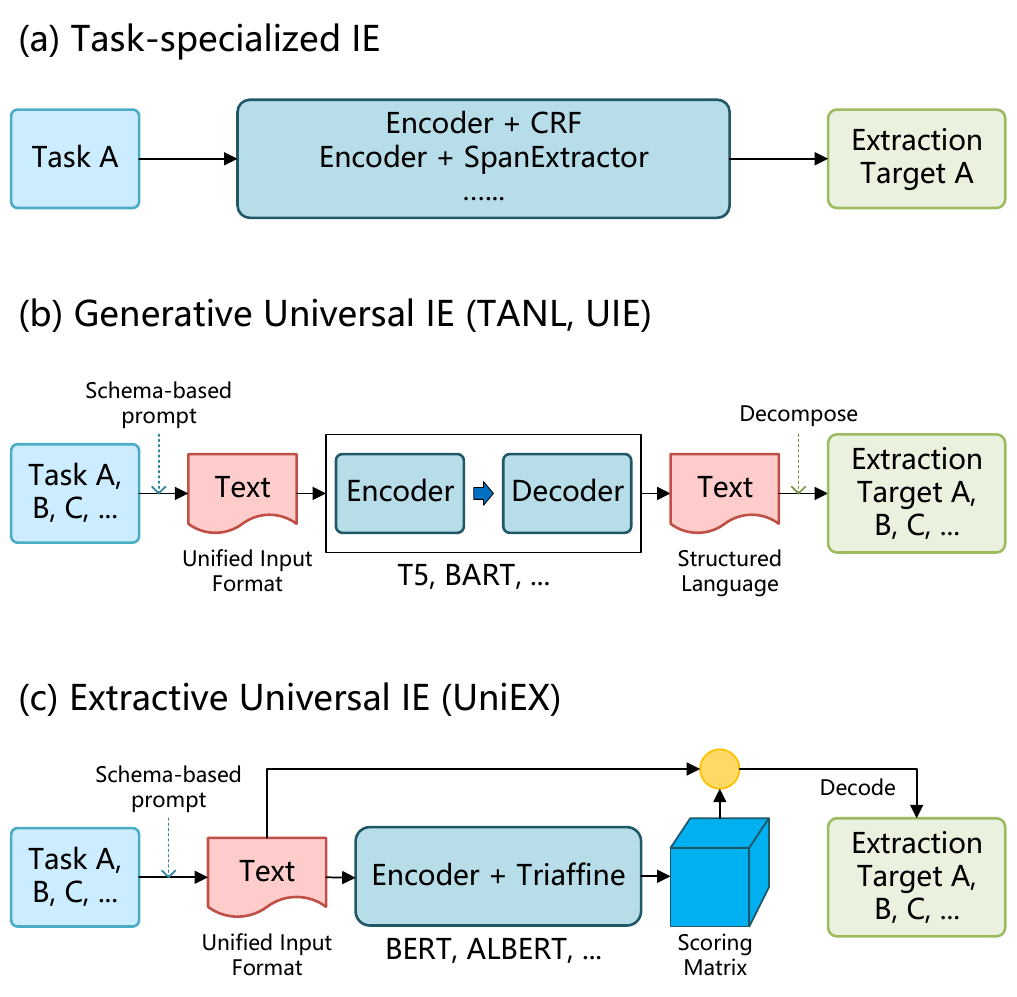}}
\vskip -0.1in
\caption{(a) Task-specific IE methods: isolated structures and schemas. (b) Typical generative universal IE: unified modeling via text or structure generation. (c) Our extractive universal IE: unified modeling via traffine attention mechanism and auto-encoder LMs.}\label{fig:Example_UIE}
\end{center}
\vskip -0.4in 
\end{figure}

Information extraction (IE) aims at automatically extracting structured information from unstructured textual sources, covering a wide range of subtasks such as named entity recognition, relation extraction, semantic role labeling, and sentiment analysis~\cite{muslea1999extraction, grishman2019twenty}. However, the variety of subtasks build the isolation zones between each other and form their own dedicated models.
Fig~\ref{fig:Example_UIE} (a) presents that the popular IE approaches handle structured extraction by the addition of task-specific layers on top of pre-trained language models (LMs) and a subsequent fine-tuning of the conjoined model~\cite{lample2016neural, luo2020hierarchical, wei2020novel, ye2022packed}. The isolated architectures and chaotic situation prevents enhancements from one task from being applied to another, which hinders the effective latent semantics sharing such as label names, and suffer from inductive bias in transfer learning~\cite{paolini2020structured}. 

With powerful capabilities in knowledge sharing and semantic generalization, large-scale LMs bring the opportunity to handle multiple IE tasks using a single framework. As shown in Fig~\ref{fig:Example_UIE} (b), by developing sophisticated schema-based prompt and structural generation specification, the IE tasks can be transformed into text-to-text and text-to-structure formats via large-scale generative LMs~\cite{dong2019unified, paolini2020structured, lu2022unified} such as T5~\cite{2020t5}. Moreover, the universal IE frameworks can learn general knowledge from multi-source prompts, which is beneficial for perceiving unseen content in low-resource scenarios. Despite their success, these generative frameworks suffer from their inherent problems, which limit their potential and performance in universal modeling. 
Firstly, the schema-based prompt and contextual information are synthetically encoded for generating the target structure, which is not conducive to directly leveraging the position information among different tokens. 
Secondly, the generative architecture utilizes the token-wise decoder to obtain the target structure, which is extremely time-consuming.

The aforementioned issues prompt us to rethink the foundation of IE tasks. Fundamentally, we discover that the extraction targets of different IE tasks involve the determination of semantic roles and semantic types, both of which can be converted into span formats by the correlation of the inside tokens in the passage. For instance, an entity type is the boundary detection and label classification of a semantic role, while a relation type can be regarded as the semantic association between specific semantic roles. From this perspective, the IE tasks can be decoded using a span-extractive framework, which can be uniformly decomposed as several atomic operations: i) Span Detection, which locates the boundaries of the mentioned semantic roles; ii) Span Classification, which recognizes the semantic types of the semantic roles; iii) Span Association, which establishes and measures the correlation between semantic roles to determine semantic types. According to the above observation, we propose a new paradigm for universal IE, called \textbf{Uni}fied \textbf{Ex}traction model (UniEX) as Figure~\ref{fig:Example_UIE} (c). Specifically, we first introduce a rule-based transformation to bridge various extraction targets and unified input formats, which leverages task-specific labels with identifiers as the schema-based prompt to learn general IE knowledge. Then, recent works~\cite{liu2019multi,yang2022zero} state that the auto-encoder LMs with bidirectional context representations are more suitable for natural language understanding. Therefore, We employ BERT-like LMs to construct an extractive architecture for underlying semantic encoding. Finally, inspired by the successful application of span-decoder and biaffine network to decode entity and relation with a scoring matrix~\cite{yu2020named, li2020unified, yuan2022fusing}, we introduce a triaffine attention mechanism for structural decoding, which jointly considers high-order interactions among multiple factors, including tasks, labels and inside tokens. Each triaffine scoring matrix is assigned to a demand-specific prompt for obtaining span-extractive objectives.

Through extensive experiments on several challenging benchmarks of $4$ main IE tasks (entity/relation/event/sentiment extraction), we demonstrate that compared with the state-of-the-art universal IE models and task-specific low-resource approaches, our UniEX achieves a substantial improvement in performance and efficiency with supervised, few-shot and zero-shot settings.

Our main contributions are summarized as:
\begin{itemize}
    \item We develop an efficient and effective universal IE paradigm by converting all IE tasks into joint span classification, detection and association problem.
    \item We introduce UniEX, a new unified extractive framework that utilizes the extractive structures to encode the underlying information and control the schema-based span decoding via the triaffine attention mechanism.
    \item We apply our approach in low-resource scenarios, and significant performance improvements suggest that our approach is potential for attaching label information to generalized objects and transfer learning. Our code will be made publicly available.
\end{itemize}

\section{Related Work}
\paragraph{Unified NLP Task Formats} Since the prompt-tuning can improve the ability of language models to learn common knowledge and fix the gap across different NLP tasks, recent studies show the necessity of unifying all NLP tasks in the format of a natural language response to natural language input~\cite{raffel2020exploring, sanh2022multitask, wei2021finetuned}. Previous unified frameworks usually cast parts of text problems as question answering~\cite{mccann2018natural} or span extraction~\cite{keskar2019unifying} tasks. TANL~\cite{paolini2020structured} frames the structured prediction tasks as a translation task between augmented natural languages. By developing a text-to-text architecture, T5~\cite{raffel2020exploring} makes prompts to effectively distinguish different tasks and provide prior knowledge for multitask learning. UIE~\cite{lu2022unified} uniformly models IE tasks with a text-to-structure framework, which encodes different extraction structures via a structured extraction language, adaptively generates varying targets via a structural schema instructor. Although effective, such methods focus on generative styles and thus cannot be adapted to the knowledge selection for vast label-based models. It motivates us to design an efficient and effective universal IE method, where we develop unified Extraction (EX) formats and triaffine attention mechanism.

\paragraph{Label Information}
Label semantics is an important information source, which carries out the related meaning induced from the data~\cite{hou2020few, ma2022label, mueller2022label}. The L-TapNet~\cite{hou2020few} introduces the collapsed dependency transfer mechanism to leverage the semantics of label names for few-shot tagging tasks. LSAP~\cite{mueller2022label} improves the generalization and data efficiency of few-shot text classification by incorporating label semantics into the pre-training and fine-tuning phases of generative LMs. Together, these successful employments of label knowledge in low-resource setting motivates us to introduce label semantics into our unified inputs to handle few-shot and zero-shot scenarios.

\section{Approaches}
Generally, there are two main challenges in universally modeling different IE tasks via the extractive architecture. Firstly, IE tasks are usually demand-driven, indicating that each pre-defined schema should correspond to the extraction of specific structural information. Secondly, due to the diversity of IE tasks, we need to resolve appropriate structural formats from the output sequence to accommodate different target structures, such as entity, relation and event. In this section, we outline how the UniEX exploits a shared underlying semantic encoder to learn the prompt and text knowledge jointly, and conduct various IE tasks in a unified text-to-structure architecture via the triaffine attention mechanism.

\begin{figure*}[t]
\begin{center}
\centerline{\includegraphics[width=1.0 \linewidth]{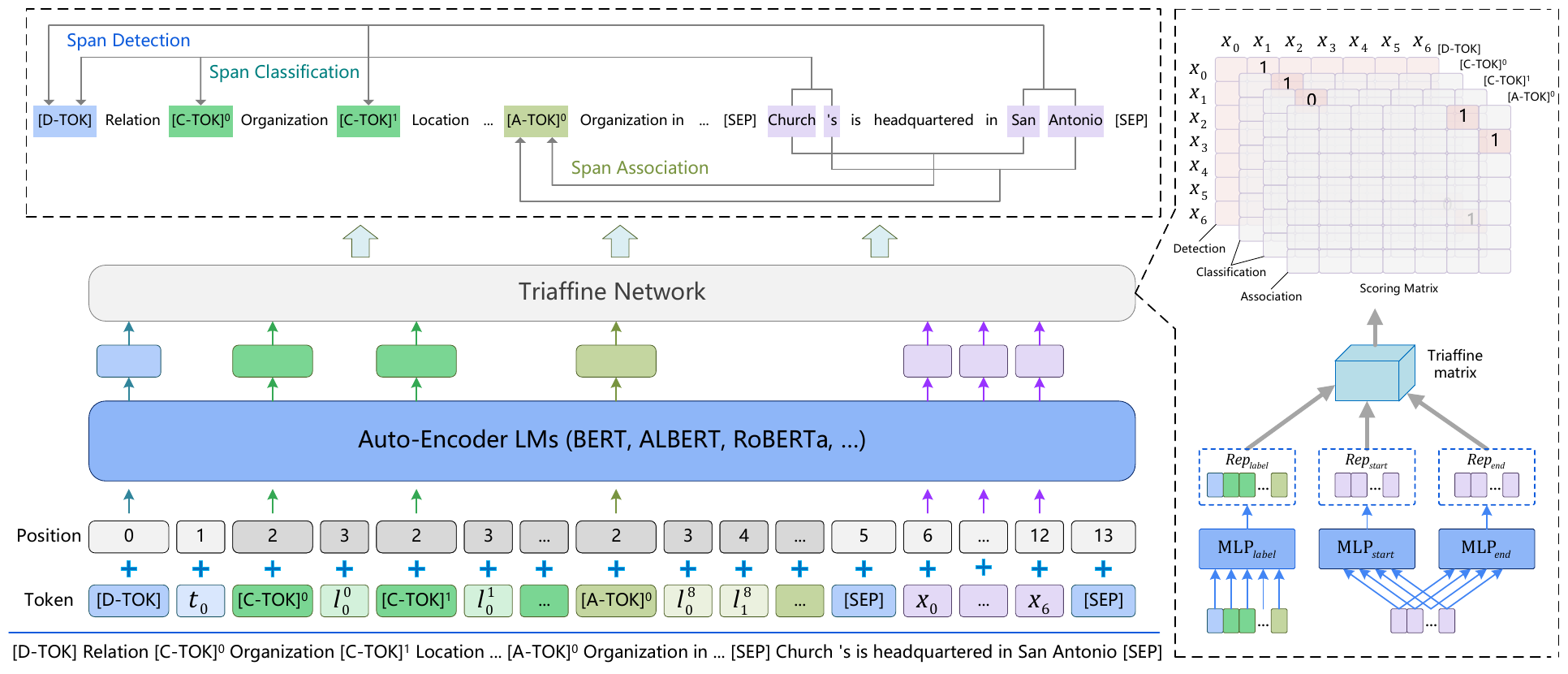}}
\vskip -0.1in
\caption{The overall architecture of UniEX. The sample text comes from CoNLL04~\cite{roth2004linear}.}
\label{fig:UniEX_Model}
\end{center}
\vskip -0.3in
\end{figure*}

\subsection{The UniEX Framework}
\subsubsection{Unified Input}
Formally, given the task-specific pre-defined schema and texts, the universal IE model needs to adaptively capture the corresponding structural information from the text indicated by the task-relevant information. To achieve this, we formulate a unified input format consisting of task-relevant schema and text, as shown in Figure~\ref{fig:UniEX_Model}. To promote the sharing of generalized knowledge across different IE tasks, we choose to simply use the task-based and label-based schemas as prompt rather than elaborate questions, fill-in blanks or structural indicators. To achieve proper prompt representation, we introduce several special tokens {\tt[D-TOK]}, {\tt[C-TOK]} and {\tt[A-TOK]} as identifiers, uniformly replacing the corresponding schema representations in the input sentence. Here, {\tt[D-TOK]} inherits the ability of {\tt[CLS]} to capture the global semantic information. {\tt[C-TOK]} and {\tt[A-TOK]} inherit the ability of {\tt[SEP]}, thus remaining to use token representation to symbolize the connotation of subsequent schemas. Consider an input set denoted as $(s, x)$, includes the following: i) task-based schema $s_d$ for span detection, ii) label-based schemas $s_c$ for span classification and $s_a$ for span association, iii) one passage $x=\{x_1, \ldots, x_{N_x}\}$. The input sentence with $N_s=N_{sd}+N_{sc}+N_{sa}$ schemas and $N_x$ inside tokens can be denoted as:
\begin{equation}
\small
\begin{aligned}
    x_{i n p}=&\left\{[\mathrm{~D}\text{-}\mathrm{TOK}]^i\ s_d^i\right\}_{i=1}^{N_{sd}}\ \left\{[\mathrm{~C}\text{-}\mathrm{TOK}]^i\ s_c^i\right\}_{i=1}^{N_{sc}}\\
&\left\{[\mathrm{~A}\text{-}\mathrm{TOK}]^i\ s_a^i\right\}_{i=1}^{N_{sa}}\ [\mathrm{SEP}]\ x\ [\mathrm{SEP}].
\end{aligned}
\end{equation}

\subsubsection{Backbone Network}
\begin{figure}[t]
\begin{center}
\centerline{\includegraphics[width=0.9 \linewidth]{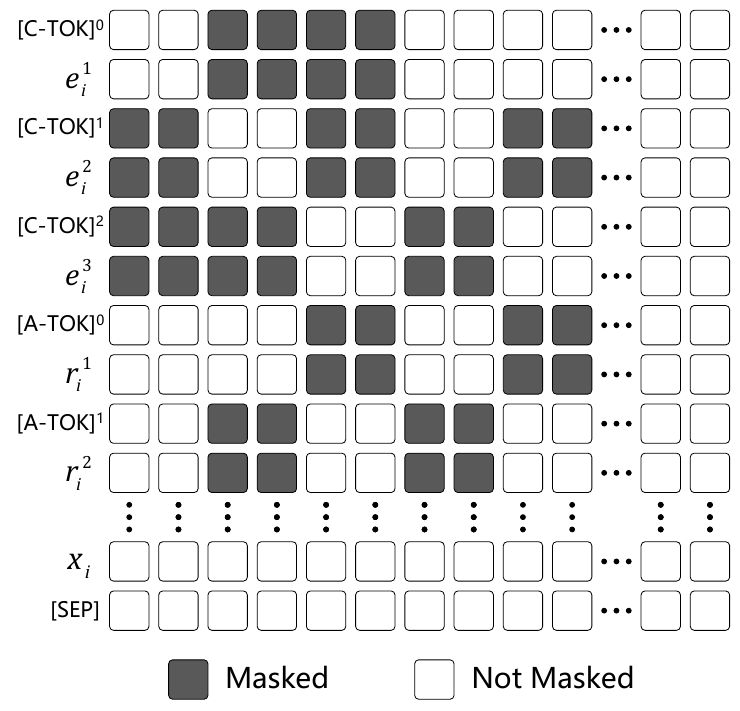}}
\vskip -0.1in
\caption{Schema-based Attention Mask Matrix of the relation extraction task with triplet type $(e^1,r^1,e^2)$ and $(e^1,r^2,e^3)$. The relation and entity types are internally invisible, whereas the paired relation and entity types can attend to each other.}\label{fig:MASK_Matrix}
\end{center}
\vskip -0.4in
\end{figure}

In our UniEX framework, we employ the BERT-like LMs as the extractive backbone, such as RoBERTa~\cite{liu2019roberta} and ALBERT~\cite{lan2020albert}, to integrate the bidirectional modeled input $x_{inp}$. Note that the unified input contains multiple labels, resulting in undesired mutual influence across different labels and leading to a misunderstanding of the correspondence between the label and its structural format during the decoding phase. Meanwhile, in some tasks, the large number of labels allows schemas to take up excessive locations, squeezing the space for text. Referring to the embedding methods in the UniMC~\cite{yang2022zero}, we address these issues from several perspectives, including position id and attention mask. Firstly, to avoid the information interference caused by the mutual interaction within label-based schemas, we constantly update the position id $pos$ to tell apart intra-information in the label. In this way, the position information of label-relevant tokens is coequally treated based on their position embedding, and the refreshed location information for the first token of each label-based schema avoids the natural increase of the location id. Then, as shown in Figure~\ref{fig:MASK_Matrix}, due to the detailed correlation among schema-based prompts in the IE tasks, we further introduce a schema-based attention mask matrix $M_{mask}$ in the self-attention calculation to control the flow of labels, ensuring that unrelated labels are invisible to each other. In particular, different entity, relation and event types are invisible to each other, while relation and event types can contact their bound entity types.

Furthermore, we take the encoded hidden vector from the last Transformer-based layer, where we combine the special tokens part as the schema representations $H_s \in \mathbb{R}^{N_s \times d}$ and the passage tokens part as the text representations $H_x \in \mathbb{R}^{N_x \times d}$ with hidden size $d$.
\begin{equation}
\small
H_s, H_x=\operatorname{Encoder}\left(x_{i n p}, pos, M_{\text {mask}}\right)
\end{equation}

\subsubsection{Triaffine Attention for Span Representation}
After obtaining the schema representations and text representations from the auto-encoder LM, the following challenge is \textit{how to construct a unified decoding format that is compatible with different IE structures, with the goal of adaptively exploiting schemas to control various extraction targets.} Take the example in Figure~\ref{fig:Example_UniEX}, for the event extraction system, we locate the start and end indices of the words boundary ``Dariues'', ``Ferguson'' and ``injure'' as the semantic roles, categorized as the \textit{Agent}, \textit{Victim} and \textit{Trigger} semantic types (entity/trigger) respectively, and collectively to the \textit{Injure} semantic type (event). For the relation extraction system, we associate the semantic roles ``Betsy Ross'' and ``Philadelphia'' by attaching their intersecting information to the \textit{Live in} semantic type (relation). In conjunction with the discussions in the  \nameref{sec:introduction}, we consider two elements for universally modeling IE tasks as joint span detection, classification and association: I) Different extraction targets are presented in the form of span, relying on unified information carriers to accommodate various semantic roles and semantic types. II) The span-extractive architecture is necessary for establishing schema-to-text information interaction, which can adaptively extract schema-related semantic information from text. 

For the first proposal, we introduce two information carriers for decoding heterogeneous IE structures in a unified span format:

\begin{figure}[t]
\begin{center}
\centerline{\includegraphics[width=1 \linewidth]{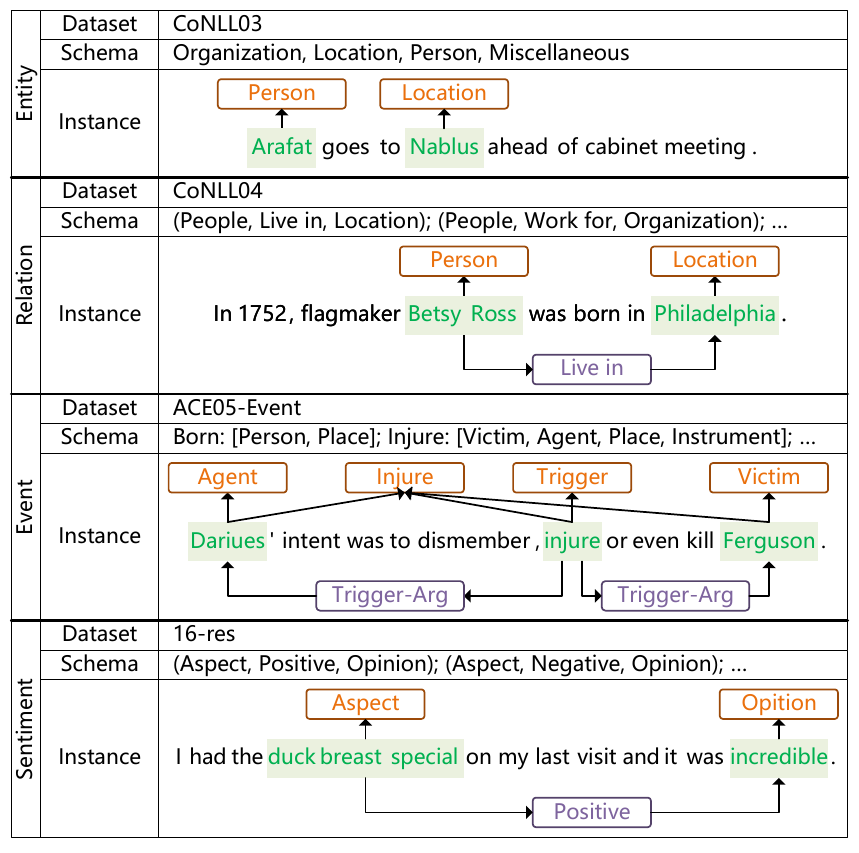}}
\vskip -0.1in
\caption{Uniformly modeling different extraction targets as joint span detection, classification and association with sampling from selected datasets.}\label{fig:Example_UniEX}
\end{center}
\vskip -0.4in
\end{figure}

{\noindent 1. \textbf{Structural Table}} indicates a rank-2 scoring matrix corresponding to a particular schema, which accommodates the semantic information required for span-extractive parsing.

{\noindent 2. \textbf{Spotting Designator}} indicates the location of spans in the preceding structural table, which represent extraction targets corresponding to the particular schema.

For the second proposal, we attempt to explore the internal interaction of the inside tokens by converting the text representation into span representation. Then, we apply two separate FFNs to create different representations ($H_x^s\ /\ H_x^e$) for the start/end positions of the inside tokens. To further interact such multiple heterogeneous factors simultaneously, we define the deep triaffine transformation with weighted matrix $\mathcal{W} \in \mathbb{R}^{d \times d \times d}$, which apply the triaffine attention to aggregate the schema-wise span representations by considering schema as queries as well as start/end of the inside tokens as keys and values. In this process, the triaffine transformation injects each schema information into the span representations and resolves the corresponding extraction targets. It creates a $N_s \times N_x \times N_x$ scoring tensor $S$ by calculating continuous matrix multiplication as following:
\begin{equation}
\small
\begin{gathered}
H_x^s=\operatorname{F F N}_s\left(H_x\right), \\
H_x^e=\operatorname{F F N}_e\left(H_x\right), \\
S=\sigma(\mathcal{W} \times_1 H_s \times_2 H_x^s \times_3 H_x^e),
\end{gathered}
\end{equation}
where $\times_k$ is the matrix multiplication between input tensor and dimension-$k$ of $\mathcal{W}$. $\sigma(*)$ denotes the Sigmoid activation function.

At this point, the tensor $S$ provides a mapping score from the schema to internal spans of the text, where each rank-2 scoring matrix corresponding to a specific schema is the structural table. For the $r$-th structural table, the affine score of each span $(p,q)$ that starts with $p$ and ends with $q$ can be denoted as $S_{r,p,q} \in [0,1]$, while the affine score of a valid span in the structural table is the spotting designator. We divide all $N_s$ structural tables into three parts according to the distribution of the schemas, among them, $N_{sd}$ for span detection, $N_{sc}$ for span classification, and $N_{sa}$ for span association. For different schemas, we develop their spotting designators by following strategies:

{\noindent \textbf{Span Detection}}: In particular, we usually use the structural table derived from the task-based schema representation for span detection, which can be obtained from the hidden state of the special token {\tt[CLS]}. Since the {\tt[CLS]} token is mutually visible to other schemas, the task-based schema representation can capture the span-related semantic information of the semantic roles from the task and label names. The spotting designators identify the start and end indices of the $i$-th semantic roles as ($s_i, e_i$) using the axes.

{\noindent \textbf{Span Classification}}: The label-based schema representations for entity/argument/trigger/event types are used for span classification. The spotting designators are identical with the span positions of the semantic roles, indicating that the semantic type of the $i$-th span can be identified by attaching to the ($s_i, e_i$) position in the corresponding structural table.

{\noindent \textbf{Span Association}}: The label-based schema representations for relation/sentiment types are used for span association. In this process, we model the potentially related semantic roles and correlate them to corresponding semantic types. The spotting designators locate at two interleaved positions associated with the semantic roles of the semantic type, that is, for the $i$-th and $j$-th spans, the extraction target is transformed to the identification of the ($s_i, s_j$) and ($e_i, e_j$) positions in the corresponding structural table.

Note that all span values in the structural table for label-based schemas are masked except for the spotting designators, because we only need to observe the semantic types and semantic association among the detected spans. Specifically, the spotting designators for span detection are the spans with $q \geq p$, and the spotting designators for span classification and association are defined by the position consistency and interleaving of valid spans with $S_{r,p,q} = 1$ in span detection.

\begin{table*}[t]
\centering
\small
\resizebox{0.95\textwidth}{!}{

\begin{tabular}{cccc|cc|cc}
\toprule
 \textbf{Task} & \textbf{Dataset} & \textbf{Domain} & \textbf{Metric} & \textbf{\makecell{TANL\\220M}} & \textbf{\makecell{UniEX\\132M}} & \textbf{\makecell{UIE\\770M}} & \textbf{\makecell{UniEX\\372M}} \\

\midrule
\multirow{4}{*}{\makecell{Entity \\Extraction}} & ACE04 & News, Speech & Entity F1 &  -  &  -  &  86.52  & \textbf{87.12} \\
&ACE05-Ent & News, Speech & Entity F1 &  84.90  &  \textbf{85.96} & 85.52  & \textbf{87.02}\\
&CoNLL03 & News  & Entity F1 &  91.70  &  92.13  &  92.17  & \textbf{92.65}\\
&GENIA &  Biology & Entity F1 &  76.40  &  \textbf{76.69}  &  -  &  -  \\

\midrule
\multirow{4}{*}{\makecell{Relation\\Extraction}} & ACE05-Rel & News, Speech & Relation Strict F1 &  \textbf{63.70} & 63.64  &  64.68  & \textbf{66.06} \\
&CoNLL04 & News & Relation Strict F1 &  71.40 &  \textbf{71.79}  &  73.07  & \textbf{73.40}\\
&SciERC & Scientific & Relation Strict F1 & - & - & 33.36  & \textbf{38.00}\\
&ADE  &   Medicine &  Relation Strict F1  &  80.60  &  83.81  &  -  &  -\\

\midrule
\multirow{4}{*}{\makecell{Event\\Extraction}} & \multirow{2}[1]{*}{ACE05-Evt} & \multirow{2}[1]{*}{News, Speech} & Event Trigger F1 &  68.40  & \textbf{70.86} &  72.63  & \textbf{74.08}\\
&      &       & Event Argument F1  & 47.60 & \textbf{50.67} &\textbf{54.67}  & 53.92\\
&\multirow{2}[1]{*}{CASIE} & \multirow{2}[1]{*}{Cybersecurity} & Event Trigger F1   & - & - & 68.98  & \textbf{71.46}\\
&      &       & Event Argument F1 & - & - & 60.37  & \textbf{62.91}\\

\midrule
\multirow{4}{*}{\makecell{Sentiment\\Extraction}} 
& 14-res & Review & Sentiment Triplet F1 &  - & -  &  73.78  & \textbf{74.77} \\
& 14-lap & Review &  Sentiment Triplet F1 &  - & -  & 63.15  & \textbf{65.23}\\
& 15-res & Review &  Sentiment Triplet F1 & - & -  & 66.10  & \textbf{68.58}\\
& 16-res &  Review & Sentiment Triplet F1 & - & -  & 73.87  & \textbf{76.02}\\

\bottomrule
\end{tabular}%
}
 \vskip -0.05in
\caption{
    Overall results of universal IE approaches on different datasets for entity/relation/event/sentiment extraction tasks. \textbf{Base} refers to TANL and UniEX respectively using T5-base and RoBERTa-base as the backbone. \textbf{Large} refers to UIE and UniEX respectively using T5-large and RoBERTa-large as the backbone.
}
\label{tab:result_supervised}
\vskip -0.1in
\end{table*}



\subsection{EX Training Procedure}
Given the input sentence $x_{inp}$, We uniformly reformat different output targets as a rank-3 matrix $Y$, sharing the same spotting designators as the triaffine scoring matrix. Similarly, we denote the value of each valid span as $Y_{r,p,q} \in \{0,1\}$, with $Y_{r,p,q}=1$ denoting the desirable span for a ground-truth and $Y_{r,p,q}=0$ denoting the meaningless span for semantic role or semantic type. Hence it is a binary classification problem and we optimize our models with binary cross-entropy:
\begin{equation}
\small
 \operatorname{BCE}(y, \hat{y}) =-(y \cdot \log (\hat{y})+(1-y) \cdot \log (1-\hat{y})), 
\end{equation}
\begin{equation}
\small
\mathcal{L} =\sum_{r=1}^{N_s} \sum_{p=1}^{N_x} \sum_{q=1}^{N_x} \operatorname{BCE}\left(Y_{r, p, q}, S_{r, p, q}\right).
\end{equation}

\section{Experiments}
To verify  the effectiveness of our UniEX, we conduct extensive experiments on different IE tasks with supervised (high-resource), few-shot and zero-shot (low-resource) scenarios.

\begin{table*}[t]
    \centering
    \setlength{\tabcolsep}{1mm}
    \resizebox{2\columnwidth}{!}{
    \begin{tabular}{lcccccccc}
    \toprule
        \multirow{3}{*}{\textbf{Models}} & \multicolumn{4}{c}{\textbf{Intra}} & \multicolumn{4}{c}{\textbf{Inter}}\\
        \cmidrule(lr){2-5} \cmidrule(lr){6-9}
        & \multicolumn{2}{c}{\textbf{1$\sim$2-shot}} & \multicolumn{2}{c}{\textbf{5$\sim$10-shot}} & \multicolumn{2}{c}{\textbf{1$\sim$2-shot}} & \multicolumn{2}{c}{\textbf{5$\sim$10-shot}}\\
        \cmidrule(lr){2-3}\cmidrule(lr){4-5} \cmidrule(lr){6-7}\cmidrule(lr){8-9}
         & 5 way & 10 way & 5 way & 10 way & 5 way & 10 way & 5 way & 10 way \\
         \cmidrule(lr){1-1}\cmidrule(lr){2-5} \cmidrule(lr){6-9}
         ProtoBERT$^{\dag}$ & 23.45\small\small{\textpm0.92} & 19.76\small{\textpm0.59} & 41.93\small{\textpm0.55} & 34.61\small{\textpm0.59} & 44.44\small{\textpm0.11} & 39.09\small{\textpm0.87} & 58.80\small{\textpm1.42} & 53.97\small{\textpm0.38} \\
         NNShot$^{\dag}$ & 31.01\small{\textpm1.21} & 21.88\small{\textpm0.23} & 35.74\small{\textpm2.36} & 27.67\small{\textpm1.06} & 54.29\small{\textpm0.40} & 46.98\small{\textpm1.96} & 50.56\small{\textpm3.33} & 50.00\small{\textpm0.36} \\
         ESD & 41.44\small{\textpm1.16} & 32.29\small{\textpm1.10} & 50.68\small{\textpm0.94} & 42.92\small{\textpm0.75} & 66.46\small{\textpm0.49} & 59.95\small{\textpm0.69} & \textbf{74.14\small{\textpm0.80}} & 67.91\small{\textpm1.41} \\
         DecomMeta & 52.04\small{\textpm0.44} & 43.50\small{\textpm0.59} & 63.23\small{\textpm0.45} & \textbf{56.84\small{\textpm0.14}} & 68.77\small{\textpm0.24} & 63.26\small{\textpm0.40} & 71.62\small{\textpm0.16} & 68.32\small{\textpm0.10} \\
         \textbf{UniEX} & \textbf{53.92\small{\textpm0.39}} & \textbf{45.67\small{\textpm0.53}} & \textbf{63.26\small{\textpm0.14}} & 56.65\small{\textpm0.27} & \textbf{69.37\small{\textpm0.19}} & \textbf{64.53\small{\textpm0.05}} & 
         73.79\small{\textpm0.32} & \textbf{69.63\small{\textpm0.45}} \\
        \bottomrule
    \end{tabular}
    }
     \vskip -0.05in
    \caption{F1 scores with standard deviations on FewNERD. $^{\dag}$ denotes the results reported from~\citet{ding2021few}.}
    \label{tab:performance_comparison_fewnerd}
\end{table*}

\begin{table*}[t]
    \centering
    \setlength{\tabcolsep}{1mm}
    \resizebox{2\columnwidth}{!}{
    \begin{tabular}{lcccccccc}
    \toprule
        \multirow{3}{*}{\textbf{Models}} & \multicolumn{4}{c}{\textbf{1-shot}} & \multicolumn{4}{c}{\textbf{5-shot}}\\
        \cmidrule(lr){2-5} \cmidrule(lr){6-9}

         & News & Wiki & Social & Mixed & News & Wiki & Social & Mixed \\
         \cmidrule(lr){1-1}\cmidrule(lr){2-5} \cmidrule(lr){6-9}
        
         TransferBERT$^{\ddag}$ & 4.75\small{\textpm1.42} & 0.57\small{\textpm0.32} & 2.71\small{\textpm0.72} & 3.46\small{\textpm0.54} & 15.36\small{\textpm2.81} & 3.62\small{\textpm0.57} & 11.08\small{\textpm0.57} & 35.49\small{\textpm7.60} \\
         Matching Network$^{\ddag}$ & 19.50\small{\textpm0.35} & 4.73\small{\textpm0.16} & 17.23\small{\textpm2.75} & 15.06\small{\textpm1.61} & 19.85\small{\textpm0.74} & 5.58\small{\textpm0.23} & 6.61\small{\textpm1.75} & 8.08\small{\textpm0.47} \\
         ProtoBERT$^{\ddag}$ & 32.49\small{\textpm2.01} & 3.89\small{\textpm0.24} & 10.68\small{\textpm1.40} & 6.67\small{\textpm0.46} & 50.06\small{\textpm1.57} & 9.54\small{\textpm0.44} & 17.26\small{\textpm2.65} & 13.59\small{\textpm1.61} \\
         L-TapNet+CDT$^{\ddag}$ & 44.30\small{\textpm3.15} & 12.04\small{\textpm0.65} & 20.80\small{\textpm1.06} & 15.17\small{\textpm1.25} & 45.35\small{\textpm2.67} & 11.65\small{\textpm2.34} & 23.30\small{\textpm2.80} & 20.95\small{\textpm2.81} \\
         DecomMeta & 46.09\small{\textpm0.44} & 17.54\small{\textpm0.98} & 25.14\small{\textpm0.24} & 34.13\small{\textpm0.92} & 58.18\small{\textpm0.87} & \textbf{31.36\small{\textpm0.91}} & 31.02\small{\textpm1.28} & 45.55\small{\textpm0.90}\\
         \textbf{UniEX} & \textbf{58.51\small{\textpm0.14}} & \textbf{18.20\small{\textpm0.45}} & \textbf{34.67\small{\textpm0.25}} & \textbf{39.28\small{\textpm0.55}} & \textbf{66.08\small{\textpm0.42}} & 29.68\small{\textpm0.32} & \textbf{38.64\small{\textpm1.29}} & \textbf{54.25\small{\textpm0.35}} \\
        \bottomrule
    \end{tabular}
    }
     \vskip -0.05in
    \caption{F1 scores with standard deviations on Cross-Dataset. $^{\ddag}$ denotes the results reported from~\citet{hou2020few}.}
    \label{tab:performance_comparison_crossdataset}
\end{table*}

\subsection{Experimental Setup}
For the supervised setting, we follow the preparation in TANL~\cite{paolini2020structured} and UIE~\cite{lu2022unified} to collect 14 publicly available IE benchmark datasets and cluster the well-representative IE tasks into 4 groups, including entity, relation, event and structured sentiment extraction. In particular, for each group, we design a corresponding conversion regulation to translate raw data into the unified EX format.

Then, for the few-shot setting, we adopt the popular datasets FewNERD~\cite{ding2021few} and Cross-Dataset~\cite{hou2020few} in few-shot entity extraction and domain partition as~\cite{ma2022decomposed}. For the zero-shot setting, we use the common zero-shot relation extraction datasets Wiki-ZSL~\cite{chen2021zs} and FewRel~\cite{han2018fewrel} and follow the same process of data and label splitting as~\cite{chia2022relationprompt}. Following the same evaluation metrics as all previous methods, we use span-based offset Micro-F1 with strict match criteria as the primary metric for performance comparison. Please refer to Appendix~\ref{sec:experiment details} for more details on dataset descriptions, unified EX input formats, metrics and training implementation.

\subsection{Experiments on Supervised Settings}
In our experiment, under the high-resource scenario, we compare our approach with the state-of-the-art generative universal IE architectures that provide a universal backbone for IE tasks based on T5~\cite{2020t5}, including TANL~\cite{paolini2020structured} and UIE~\cite{lu2022unified}. For a fair comparison, We only consider results without exploiting large-scale contexts and external knowledge beyond the dataset-specific information, and present the average outcomes if the baseline is conducted in multiple runs. The main results of UniEX and other baselines on 14 IE datasets are shown in Table~\ref{tab:result_supervised}. We can observe that: 1) By modeling IE as joint span detection, classification and association, and encoding the schema-based prompt and input texts with the triaffine attention mechanism, UniEX provides an effective universal extractive backbone for all IE tasks. The UniEX outperforms the universal IE models with approximate backbone sizes, achieving new state-of-the-art performance on almost all tasks and datasets. 2) The introduction of label-based schema facilitates the model learning task-relevant knowledge, while the triaffine scoring matrix establishes the correspondence between each schema and extraction targets. Obviously, the UniEX can better capture and share label semantics than using generative structures to encode underlying information. Meanwhile, triaffine transformation is a unified and cross-task adaptive operation, precisely controlling where to detect and which to associate in all IE tasks. Compared with the TANL and UIE, our approach achieves significant performance improvement on most datasets, with nearly $1.36\%$ and $1.52\%$ F1 on average, respectively.

\subsection{Experiments on Low-resource Scenarios}
To verify the generalization and transferability of UniEX in low-resource scenarios, we evaluate models under few-shot and zero-shot settings, respectively. In order to reduce the influence of noise caused by random sampling on the experiment results, we repeat the data/label selection processes for five different random seeds and report the averaged experiment results as previous works~\cite{hou2020few, chia2022relationprompt}. We use the BERT-base~\cite{devlin2019bert} as the UniEX backbone to align with other low-resource results.

Firstly, we compare the UniEX with the competitive few-shot entity extraction models. For FewNERD, we compare the proposed approach to DecomMeta~\cite{ma2022decomposed}, ESD~\cite{wang2022enhanced}, and methods from~\cite{ding2021few}, e.g., ProtoBERT, NNShot. For Cross-Dataset, we compare the UniEX to DecomMeta~\cite{ma2022decomposed} and baselines reported by~\cite{hou2020few}, e.g., TransferBERT, Matching Network, ProtoBERT and L-TapNet+CDT.

Table~\ref{tab:performance_comparison_fewnerd} and ~\ref{tab:performance_comparison_crossdataset} illustrates the main results on FewNERD and Cross-Dataset of our approach alongside those reported by previous methods. It can be seen that UniEX achieves the best performance under different type granularity and domain divisions, and outperforms the prior methods with a large margin. Compare with DecomMeta on Cross-Dataset, UniEX achieves a performance improvement up to 6.94\% and 5.63\% F1 scores on average in 1-shot and 5-shot, which demonstrates the effectiveness of our approach in learning general IE knowledge. It indicates that even without pre-training on large-scale corpus, our approach can still sufficiently excavate the semantic information related with objective entities from label names, which enhances the understanding of task-specific information when data is extremely scarce.

Secondly, we compare UniEX with the latest baselines TableSequence~\cite{wang2020two} and RelationPrompt~\cite{chia2022relationprompt} on zero-shot relation triplet extraction task for Wiki-ZSL and Few-Rel datasets in Table~\ref{tab:result_zeroshot}. In both single-triplet and multi-triplet evaluation, UniEX consistently outperforms the baseline models in terms of Accuracy and overall F1 score respectively, which demonstrates the ability of our approach to handle unseen labels. Although we observe a lack of advantage in recall score for multi-triplet evaluation, the significant improvement in precision allowed our approach to achieve a balanced precision-recall ratio. The reason for such difference is probably because the directional matching in the triaffine transformation will tend to guide the model to predict more credible targets.

\begin{table}[t]
	\small
	\renewcommand{\arraystretch}{1.1}
	\setlength\tabcolsep{2.5pt}
	\centering
\resizebox{0.48\textwidth}{20mm}{		
  \begin{tabular}{@{\extracolsep{\fill}}cccccc@{}}
			\toprule
			\multirow{2}{*}{\textbf{Dataset}}& \multirow{2}{*}{\textbf{Model}} & \textbf{Single-Triplet} & \multicolumn{3}{c}{\textbf{Multi-Triplet}} \\
			\cmidrule{3-6}
			& & \textit{Acc.} & \textit{P.} & \textit{R.} & \textit{F1}\\
			\midrule
			\multirow{3}{*}{Wiki-ZSL}& TableSequence & 14.47 & 43.68 & 3.51 & 6.29\\
            & RelationPrompt & 16.64 & 29.11 & \textbf{31.00} & 30.01\\
            & UniEX & \textbf{26.84} & \textbf{58.22} & 25.85 & \textbf{34.94}\\
			\midrule
		    \multirow{3}{*}{FewRel}& TableSequence & 11.82 & 15.23 & 1.91 & 3.40\\
            & RelationPrompt & 22.27 & 20.80 & \textbf{24.32} & 22.34\\
            & UniEX & \textbf{27.30} & \textbf{44.46} & 15.72 & \textbf{23.13}\\
			\bottomrule
	\end{tabular}}
  \vskip -0.05in
 	\caption{Result for zero-shot relation triplet extraction under the setting of unseen label set size $m=5$. We use the Micro-F1, Precision (P.) and Recall (R.) to evaluate the multiple triplet extraction. Evaluating single triplet extraction involves only one possible triplet for each sentence, hence we only use the Accuracy (Acc.) metric.  \label{tab:result_zeroshot}}
\end{table}

\begin{table}[t]
    \vskip -0.1in
	\small
	\renewcommand{\arraystretch}{1.1}
	\setlength\tabcolsep{1.5pt}
	\centering
		\begin{tabular}{@{\extracolsep{\fill}}cccccc@{}}
			\toprule
			\textbf{Dataset}& \textbf{CoNLL03} & \textbf{CoNLL04} & \multicolumn{2}{c}{\textbf{CASIE}} & \textbf{16-res} \\ \midrule
			\textbf{F1}& \textbf{Ent} & \textbf{Rel-S} & \textbf{Evt-Tri} & \textbf{Evt-Arg} & \textbf{Rel-S}\\
			\midrule
			W/O SAM & 28.47 & 0 & 4.03 & 0 & 0\\
            W/O TriA & 58.58 & 49.40 & 6.97 & 1.51 & 29.77\\
            W/O Label & 92.59 & 70.94 & 71.18 & 62.29 & 74.64\\ \midrule
            \textbf{UniEX} & 92.65 & 73.40 & 71.46 & 62.91 & 76.02\\ 
			\bottomrule
	\end{tabular}
 \vskip -0.05in
 	\caption{Experiment results of UniEX with different ablation strategies on the test set of four downstream datasets: CoNLL03 (entity), CoNLL04 (relation), CASIE (event) and 16-res (sentiment). \label{tab:result_ablation}}
  	\vskip -0.1in
\end{table}

\begin{table}[t]
	\small
	\renewcommand{\arraystretch}{1.1}
	\setlength\tabcolsep{4pt}
	\centering
		\begin{tabular}{@{\extracolsep{\fill}}p{1cm}<{\centering}p{1.4cm}<{\centering}p{1.4cm}<{\centering}p{1.4cm}<{\centering}p{1.4cm}<{\centering}@{}}
			\toprule
			\multirow{2}{*}{\textbf{Model}}     & \textbf{CoNLL03} & \textbf{CoNLL04} & \textbf{CASIE} & \textbf{16-res} \\
    &(sent/s) & (sent/s) & (sent/s) & (sent/s) \\
   			\midrule
			UIE& 2.1(×1.0) & 1.0(×1.0) & 1.1(×1.0) & 1.4(×1.0)\\
			\textbf{UniEX} & 16.5(×7.9) & 16.6(×16.6) & 14.9(×13.5) & 19.7(×14.1)\\
			\bottomrule
	\end{tabular}
  \vskip -0.05in
 	\caption{The efficiency comparison of UIE and UniEX with batch\_size=1. $(\times k)$ is the relative inference-speed. \label{tab:result_efficiency}}
	\vskip -0.3in
\end{table}

\subsection{Ablation Study}
\label{sec:ablation Study}
In this section, we intend to verify the necessity of key components of the UniEX, including the flow controlling and triaffine transformation. Table~\ref{tab:result_ablation} shows ablation experiment results of UniEX on four downstream tasks.

{\noindent \textbf{W/O SAM}}: removing the schema-based attention mask matrix that controls the flowing of labels. We find that model performance is almost zero on many tasks, which demonstrates the importance of eliminating intra-information of labels. AMM makes the labels unreachable to each other, effectively avoiding the mutual interference of label semantics.

{\noindent \textbf{W/O TriA}}: replacing the triaffine transformation with the multi-head selection network, which multiplies the schema and the head-to-tail span of the text respectively, and then replicates and adds them to get the scoring matrix. The significant performance decline demonstrates the important role of triaffine attention mechanism in establishing dense correspondence between schemas and text spans.

{\noindent \textbf{W/O Label}}: replacing the label names with the special token {\tt[unused n]}, which eliminates label semantics while allowing the model to still distinguish between different labels. We find a slight degradation of model performance in small datasets CoNLL03 and 16-res, indicating that the prior knowledge provided by label names can effectively compensate for the deficiency of training data. As the correspondence between schema and extraction targets is not affected, model performance in large datasets tends to stabilize.

\subsection{Efficiency Analysis}
To verify the computation efficiency of our approach on universal IE, we compare inference-speed with UIE~\cite{lu2022unified} on the four standard datasets mentioned in section \ref{sec:ablation Study}. As shown in Table~\ref{tab:result_efficiency}, we can find that since generating the target structure is a token-wise process, the inference-speed of UIE is slow and limited by the length of the target structure. On the contrary, UniEX can decode all the target structures at once from the scoring matrices obtained by triaffine transformation, with an average speedup ratio of 13.3 to UIE.

\section{Conclusion}
In this paper, we introduce a new paradigm for universal IE by converting all IE tasks into joint span detection, classification and association problems with a unified extractive framework. UniEX collaboratively learns the generalized knowledge from schema-based prompts and controls the correspondence between schema and extraction targets via the triaffine attention mechanism. Experiments on both supervised setting and low-resource scenarios verify the transferability and effectiveness of our approaches.

\section*{Limitations}
In this paper, our main contribution is an effective and efficient framework for universal IE. We aim to introduce a new unified IE paradigm with extractive structures and triaffine attention mechanism, which can achieve better performance in a variety of tasks and scenarios with more efficient inference-speed. However, it is non-trivial to decide whether a sophisticated and artificial prompt is required for complex datasets and large label sets. In addition, we only compare with limited baselines with specific datasets configurations when analyzing the performance of the UniEX in supervised, few-shot and zero-shot settings. In experiments, we implement only a few comparative
experiments between BERT~\cite{devlin2019bert} and RoBERTa~\cite{liu2019roberta} due to the limit of computational resources.

\section*{Ethical Considerations}
As an important domain of natural language processing, information extraction is a common technology in our society. It is necessary to discuss the ethical influence when using the extraction models~\cite{leidner2017ethical}. In this work, We develop a new universal IE framework, which enhances the generalization ability in various scenarios. As discussed~\citep{schramowski2019bert, schramowski2022large, blodgett2020language}, pre-trained LMs might contain human-made biases, which might be embedded in both the parameters and outputs of the open-source models. In addition, we note the potential abuse of universal IE models, as these models achieve excellent performance in various domains and settings after adapting to pre-training on large-scale IE datasets, which allows the models to be integrated into applications often without justification. We encourage open debating on its utilization, such as the task selection and the deployment, hoping to reduce the chance of any misconduct.

{\small
\bibliography{custom}
\bibliographystyle{acl_natbib}
}

\clearpage

\appendix

\section{Experiment Details}
\label{sec:experiment details}

This section describes the details of experiments, including the dataset descriptions, unified EX input formats, metrics and training implementation.

\subsection{Details of Downstream Tasks}
\subsubsection{Supervised Setting}
For the supervised setting, We conduct downstream tasks on 4 IE tasks, 14 datasets, and the detailed statistic of each dataset is shown in Table~\ref{tab:details_datasets}.

\paragraph{Entity}
We conduct entity extraction experiments on four datasets, including the flat entity dataset extraction dataset CONLL03~\cite{sang2003introduction}, and nested entity extractions datasets ACE04~\cite{doddington2004automatic}, ACE05-Ent~\cite{ace2005} and GENIA~\cite{ohta2002genia}. For the CONLL03, ACE04 and ACE05-Ent, We use the same processing and splits as~\cite{li2020unified}. For the GENIA, we follow the pre-processing steps and data split as~\cite{finkel2009nested}.

\paragraph{Relation}
We conduct relation extraction experiments on five joint entity-relation extraction datasets across several languages and domains, including CONLL04~\cite{roth2004linear}, ACE05-Rel~\cite{ace2005}, NYT~\cite{riedel2010modeling}, SciERC~\cite{luan2018multi} and ADE~\cite{gurulingappa2012development}. We follow the pre-processing versions and data split of previous works~\cite{gupta2016table, yu2020joint, luan2019general}.

\paragraph{Event}
For ACE05-Evt, we follow the same types, data splits, and pre-processing steps as~\cite{lin2020joint}. For CASIE~\cite{satyapanich2020casie}, we remove three incomplete annotated documents, then split the remaining documents into three sets as~\cite{lu2022unified}.

\paragraph{Sentiment}
We conduct sentiment extraction experiments on the sentiment triplet extraction~\cite{xu2020position} of SemEval 14/15/16 aspect sentiment analysis datasets. We employ the pre-processing datasets of the previous work~\cite{yan2021unified}.

\begin{table}[t]
  \centering
  \resizebox{0.49\textwidth}{!}{
    \begin{tabular}{c|ccc|ccc}
    \toprule
          & |Ent| & |Rel| & |Evt| & \#Train & \#Val & \#Test \\
    \midrule
    ACE04 & 7     & -     & -     & 6,202  & 745   & 812  \\
    ACE05-Ent & 7     & -     & -     & 7,299  & 971   & 1,060  \\
    CoNLL03 & 4     & -     & -     & 14,041  & 3,250  & 3,453  \\
    GENIA & 5   &   -   &   -       & 14,824 &  1,855  & 1,854  \\
    ACE05-Rel & 7     & 6     & -     & 10,051  & 2,420  & 2,050  \\
    CoNLL04 & 4     & 5     & -     & 922   & 231   & 288  \\
    NYT   & 3     & 24    & -     & 56,196  & 5,000  & 5,000  \\
    SciERC & 6     & 7     & -     & 1,861  & 275   & 551  \\
    ADE   &  2      &   1   & -    & 3,417   & 427   & 428  \\
    ACE05-Evt & -     & -     & 33    & 19,216  & 901   & 676  \\
    CASIE & 21     & -     & 5     & 11,189  & 1,778  & 3,208  \\
    14res & 2     & 3     & -     & 1,266  & 310   & 492  \\
    14lap & 2     & 3     & -     & 906   & 219   & 328  \\
    15res & 2     & 3     & -     & 605   & 148   & 322  \\
    16res & 2     & 3     & -     & 857   & 210   & 326  \\
    \bottomrule
    \end{tabular}%
  }
  \caption{
    Detailed datasets statistics.
    |*| indicates the number of categories, and \# is the number of sentences in the specific subset.
    We take sentiment types as special relation type: positive, negative, and neutral; and each sentiment triplet holds a aspect and a opinion.
}
  \label{tab:details_datasets}
\end{table}

\begin{table*}[t]
	\small
	\renewcommand{\arraystretch}{1.1}
	\setlength\tabcolsep{4pt}
	\centering
		\begin{tabular}{@{\extracolsep{\fill}}ccccc|cccc@{}}
			\toprule
			\multicolumn{1}{c}{\textbf{Backbone}} & \multicolumn{4}{c|}{\textbf{RoBERTa-large/RoBERTa-base}} & \multicolumn{4}{c}{\textbf{BERT-base}} \\
   			\hline
            \textbf{Task} &\textbf{Entity}&\textbf{Relation}&\textbf{Event}&\textbf{Sentiment}&\multicolumn{2}{c}{\textbf{Cross Dataset}}&\textbf{Wiki-ZSL}&\textbf{FewRel}\\
            	\hline
        \textbf{Phase} & \textbf{finetuning} & \textbf{finetuning} & \textbf{finetuning} & \textbf{finetuning} & \textbf{pretraining} & \textbf{finetuning} & \textbf{pretraining} & \textbf{pretraining}\\
        \hline
        \textbf{Learning Rate} & 2E-5 & 2E-5 & 2E-5 & 2E-5 & 2E-5 & 2E-5 & 2E-05 & 2E-5 \\
        \textbf{Batch Size} & 32 & 32 & 32 & 32 & 32 & 2 & 32 & 32 \\
        \textbf{Schedule} & linear & linear & linear & linear & linear & linear & linear & linear \\
        \textbf{Warmup Rate} & 0.06 & 0.06 & 0.06 & 0.06 & 0.06 & 0.06 & 0.06 & 0.06\\
        \textbf{Epoch} & 200 & 400 & 200 & 200 & 100 & 100 & 4 & 4\\
			\bottomrule
	\end{tabular}
 \caption{Hyper-parameters for UniEX-base and UniEX-large on different tasks and datasets. \label{tab:hyper-parameters}}
 \vskip -0.1in
\end{table*}

\subsubsection{Few-shot Setting}
For the few-shot setting, we conduct downstream tasks on 2 few-shot named entity recognition datasets:

\paragraph{Few-NERD}~\cite{ding2021few}. It is annotated with a hierarchy of 8 coarse-grained and 66 finegrained entity types. Two tasks are considered
on this dataset: i) Intra, where all entities in train/dev/test splits belong to different coarsegrained types. ii) Inter, where train/dev/test splits may share coarse-grained types while keeping the fine-grained entity types mutually disjoint.

\paragraph{Cross-Dataset}~\cite{hou2020few}. Four datasets focusing on four domains are used here: CoNLL2003~\cite{sang2003introduction} (news),
GUM~\cite{zeldes2017gum} (Wiki) , WNUT-2017~\cite{derczynski2017results} (social), and Ontonotes~\cite{pradhan2013towards} (mixed). Among them, we take two domains for training, one for validation, and the remaining for
test.

\subsubsection{Zero-shot Setting}
For the zero-shot setting, we conduct downstream tasks on 2 zero-shot named entity recognition datasets:

\paragraph{FewRel}~\cite{han2018fewrel} is hand-annotated for few-shot relation extraction, we further made it suitable for the zero-shot setting after data splitting into disjoint relation label sets for training, validation and testing as~\cite{chia2022relationprompt}. 

\paragraph{Wiki-ZSL}~\cite{chen2021zs} is constructed through distant supervision over Wikipedia articles and the Wikidata knowledge base.

To partition the data into seen and unseen label sets, we follow the same process as~\cite{chia2022relationprompt} to be consistent. For each dataset, a fixed number of labels are randomly selected as unseen labels while the remaining labels are treated as seen labels during training. The unseen label
set size is set to m=5 in our experiments. In order to reduce the effect of experimental noise, the label selection process is repeated for five different random seeds to produce different data folds. For each data fold, the test set consists of the sentences containing unseen labels. Five 
validation labels from the seen labels are used to select sentences for early stopping and hyperparameter tuning. The remaining sentences are treated as the train set. Hence, the zero-shot setting ensures that train, validation and test sentences belong to disjoint label sets.

\begin{table}[t]
  \centering
    \setlength{\belowcaptionskip}{-0.3cm}
  \resizebox{0.45\textwidth}{!}{
    \begin{tabular}{ccc}
    \toprule
    \textbf{Hyper-parameter} & \textbf{UniEX-base} & \textbf{UniEX-large} \\
    \midrule
    Backbone        & Roberta-large & Roberta-base \\
    Layers of Encoder & 12    & 24 \\
    Hidden Dimension & 768   & 1,024 \\
    FF hidden size & 3072 & 4096 \\
    Layer Normalize $\epsilon$ & 1e-5  & 1e-5 \\
    Attention head & 12    & 16 \\
    \bottomrule
    \end{tabular}%
  }
  \caption{Model architectures.}
  \label{tab:model-architectures}%
\end{table}%

\subsection{Evaluation Metric}
We use span-based offset Micro-F1 as the primary metric to evaluate the model as~\cite{lu2022unified}

\begin{itemize}[nosep,leftmargin=*]
\item \textbf{Entity}: an entity mention is correct if its offsets and type match a reference entity.
\item \textbf{Relation Strict}: relation with strict match, a relation is correct if its relation type is correct and the offsets and entity types of the related entity mentions are correct.
\item \textbf{Relation Triplet}: relation with boundary match, a relation is correct if its relation type is correct and the string of the subject/object are correct.
\item \textbf{Event Trigger}: an event trigger is correct if its offsets and event type matches a reference trigger.
\item \textbf{Event Argument}: an event argument is correct if its offsets, role type, and event type match a reference argument mention.
\item \textbf{Sentiment Triplet}: a correct triplet requires the offsets boundary of the target, the offsets boundary of the opinion span, and the target sentiment polarity to be all correct at the same time.
\end{itemize}

\subsection{Training Implementation}
To make a fair comparison, we first initialize UniEX-base and UniEX-large with RoBERTa-base and RoBERTa-large checkpoints~\cite{liu2019roberta} for the supervised setting, and use the BERT-base checkpoint~\cite{devlin2019bert} as the backbone for the few-shot and zero-shot settings. The model architectures are shown in Table~\ref{tab:model-architectures}. We employ Adam optimizer~\cite{kingma2015adam} as the optimizer with 1e-8 weight decay. Table~\ref{tab:hyper-parameters} shows the detailed hyper-parameters for
downstream tasks. We truncate the concatenated overall length
of schema-based prompt $s$ and raw text $x$ to 512 during training.

\begin{table}[t]
\centering
\small
\resizebox{0.45\textwidth}{!}{

\begin{tabular}{cc|cc}
\toprule
 \textbf{Task} & \textbf{Dataset} & \textbf{\makecell{UIE\\770M}} & \textbf{\makecell{UniEX\\372M}} \\

\midrule
\multirow{3}{*}{\makecell{Entity \\Extraction}} & ACE04  &   1.23	&  18.29 \\
&ACE05-Ent & 1.62 &	18.16\\
&CoNLL03 & 2.06	& 16.45\\

\midrule
\multirow{3}{*}{\makecell{Relation\\Extraction}} & ACE05-Rel & 1.64	 &  18.69 \\
&CoNLL04 & 1.00 &	16.60 \\
&SciERC & 1.02	&  17.09\\

\midrule
\multirow{2}{*}{\makecell{Event\\Extraction}} & ACE05-Evt & 1.55 & 12.93  \\
& CASIE & 1.55 & 12.93\\

\midrule
\multirow{4}{*}{\makecell{Sentiment\\Extraction}} 
& 14-res & 1.45	& 18.60 \\
& 14-lap & 1.49	& 19.78 \\
& 15-res & 1.41	& 18.37\\
& 16-res &  1.38& 19.71\\

\bottomrule
\end{tabular}%
}
 \vskip -0.05in
\caption{
    The average number of sentences generated per second by UIE and UniEX in the decoding phase.
}
\label{tab:decoding_efficiency}
\vskip -0.1in
\end{table}

\subsection{Unified Input}
Inspired by template examples in UIE~\cite{lu2022unified}, we
design a simple rule to transform the original text to a unified
EX format. In addition, we present four examples for different tasks:

An example of CONLL03 (Entity Extraction):
{\noindent \textbf{Raw text}}: \{$x$: ``Arafat goes to Nablus ahead of cabinet meeting .'', \textit{entity type}: [Location, Organization, Person, Miscellaneous], \textit{extraction target}: [(Arafat, 1, 1, Person), (Nablus, 4, 4, Location)]\}

{\noindent \textbf{Transformed Input}}: {\tt[CLS]} Entity Extraction {\tt[R-LEP]$^1$} Location {\tt[R-LEP]$^2$} Organization {\tt[R-LEP]$^3$} Person {\tt[R-LEP]$^4$} Miscellaneous {\tt[SEP]} Arafat goes to Nablus ahead of cabinet meeting . {\tt[SEP]}

An example of CONLL04 (Relation Extraction):
{\noindent \textbf{Raw text}}: \{$x$: ``In 1752 , flagmaker Betsy Ross was born in Philadelphia .'', \textit{entity-relation type}: [(Organization, organization based in, Location), (Location, location in, Location), (Person, live in, Location), (Person, work for, Organization), (Person, kill, Person)], \textit{extraction target}: [(Betsy Ross, 5, 6, Person), (Philadelphia, 10, 10, Location), (Betsy Ross, live in, Philadelphia)]\}

{\noindent \textbf{Transformed Input}}: {\tt[CLS]} Relation Extraction {\tt[R-LEP]$^1$} Location {\tt[R-LEP]$^2$} Organization {\tt[R-LEP]$^3$} Person {\tt[R-LEP]$^4$} Miscellaneous {\tt[R-LEP]$^5$} work for {\tt[R-LEP]$^6$} organization based in {\tt[R-LEP]$^7$} location in {\tt[R-LEP]$^8$} live in {\tt[R-LEP]$^9$} kill {\tt[SEP]} In 1752 , flagmaker Betsy Ross was born in Philadelphia . {\tt[SEP]}

An example of ACE05-Evt (Event Extraction):
{\noindent \textbf{Raw text}}: \{$x$: ``Sergeant Chuck Hagel was seriously wounded twice in Vietnam .'', \textit{event-trigger-argument type}: [(Born, Trigger, Person, Place), (Injure, Trigger, Victim, Agent, Place, Instrument), (Convict, Trigger, Defendant, Adjudicator, Place), ...], \textit{extraction target}: [(Chuck Hagel, 2, 3, Victim), (wounded, 6, 6, Trigger), (Vietnam, 9, 9, Place), (Injure, wounded, Chuck Hagel, Vietnam)]\}

{\noindent \textbf{Transformed Input}}: {\tt[CLS]} Event Extraction {\tt[R-LEP]$^1$} Trigger {\tt[R-LEP]$^2$} Person {\tt[R-LEP]$^3$} Place ... {\tt[R-LEP]$^i$} Born {\tt[R-LEP]$^{i+1}$} Injure ... {\tt[R-LEP]$^{n}$} Trigger-Argument {\tt[SEP]} Sergeant Chuck Hagel was seriously wounded twice in Vietnam . {\tt[SEP]}

An example of 16-res (Sentiment Extraction):
{\noindent \textbf{Raw text}}: \{$x$: ``I had the duck breast special on my last visit and it was incredible .'', \textit{entity-relation-entity type}: [(Aspect, Positive, Opinion), (Aspect, Negative, Opinion), (Aspect, Neutral, Opinion)], \textit{extraction target}: [(duck breast special, 4, 6, Aspect), (incredible, 14, 14, Opinion), (Positive, duck breast special, incredible)]\}

{\noindent \textbf{Transformed Input}}: {\tt[CLS]} Sentiment Extraction {\tt[R-LEP]$^1$} Aspect {\tt[R-LEP]$^2$} Opinion {\tt[R-LEP]$^3$} Positive {\tt[R-LEP]$^4$} Negative {\tt[R-LEP]$^5$} Neutral {\tt[SEP]} I had the duck breast special on my last visit and it was incredible . {\tt[SEP]}

\subsection{Unified Decoding}
\begin{figure*}[t]
\begin{center}
\centerline{\includegraphics[width=1.0 \linewidth]{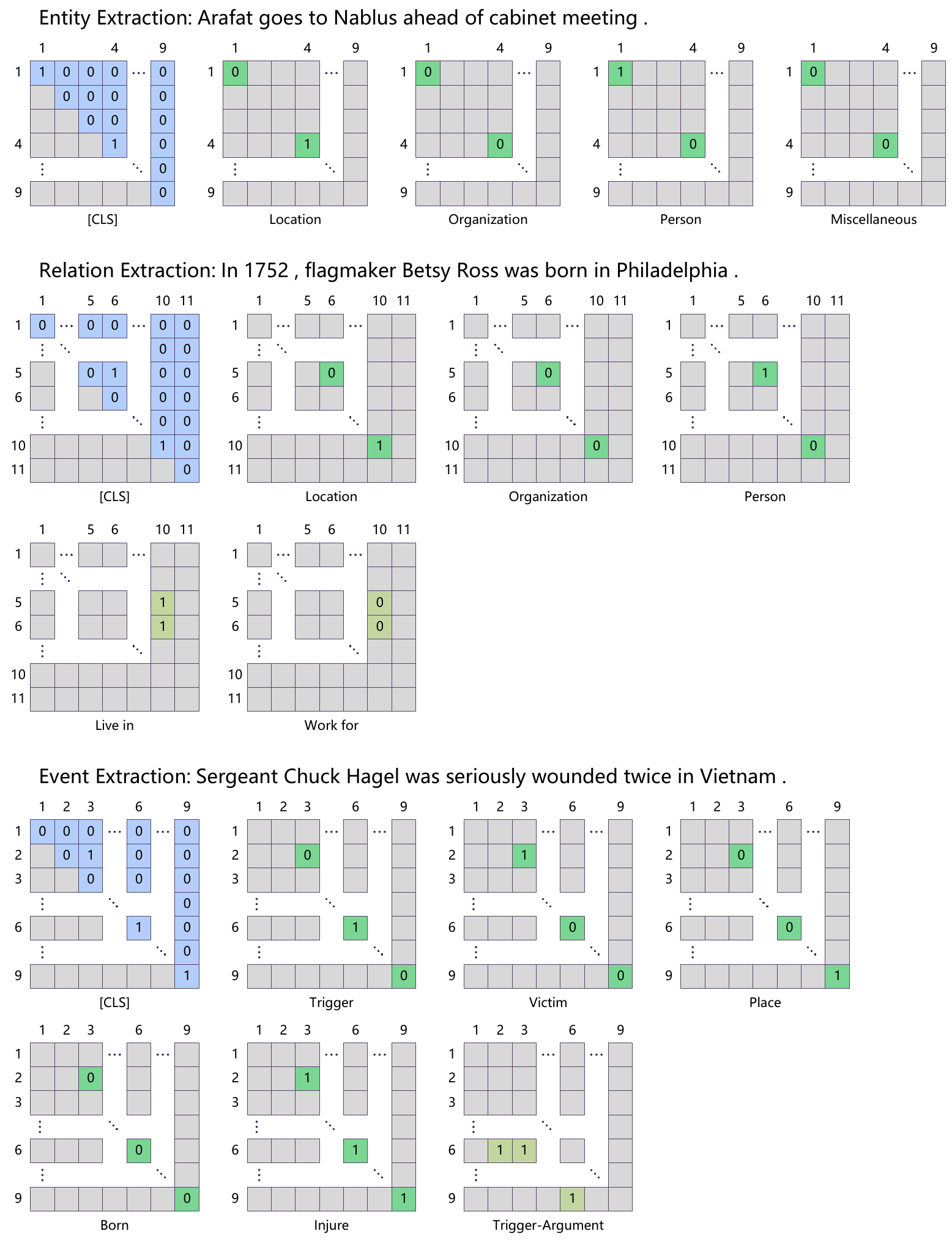}}
\vskip -0.1in
\caption{The decoding process of the UniEX. Each schema corresponds to a structural table, and each rectangle in the structural table represents an internal span, the gray spans are the invalid spans that and do not participate in model training. Other spans are spotting designators, among them, water-blue spans for span detection, viridis spans for span classification and atrovirens spans for span association.}\label{fig:UniEX_Decoding}
\end{center}
\end{figure*}

As shown in Figure~\ref{fig:UniEX_Decoding}, in order to depict the training and inference processes in more detail, we show the structural tables and spotting designators of the examples in figure~\ref{fig:Example_UniEX} in the entity/relation/event extraction tasks.

\subsection{Decoding Efficiency}
As shown in Figure~\ref{tab:decoding_efficiency}, to explicitly compare the structural decoding efficiency of different universal IE models, we illustrate the average number of sentences generated per second by UIE and UniEX during the decoding phase.

\end{document}